\documentclass[conference]{IEEEtran}
\usepackage{blindtext, graphicx}
\usepackage{amsmath}
\usepackage{amssymb}
\usepackage{marvosym}
\usepackage{multirow}
\usepackage{algorithmic}
\usepackage{algorithm}

\newcommand{\ie}{i.e.}
\newcommand{\etal}{\textit{et al.~}}

\ifCLASSINFOpdf
\else
\fi
\begin{document}
%
% paper title
% can use linebreaks \\ within to get better formatting as desired
\title{Face Recognition via Locality Constrained Low Rank Representation and Dictionary Learning}

% author names and affiliations
% use a multiple column layout for up to three different
% affiliations
\author{He-Feng Yin$^{1,2}$ ~~Xiao-Jun Wu$^{1,2}$ ~~Josef Kittler$^3$  \\
$^1$School of IoT Engineering, Jiangnan University, Wuxi 214122, China\\
$^2$Jiangsu Provincial Engineering Laboratory of Pattern Recognition and Computational Intelligence, Wuxi, China\\
$^3$Centre for Vision, Speech and Signal Processing, University of Surrey, Guildford GU2 7XH, UK\\
{\tt\small yinhefeng@126.com, wu\_xiaojun@jiangnan.edu.cn, j.kittler@surrey.ac.uk}
% For a paper whose authors are all at the same institution,
% omit the following lines up until the closing ``}''.
% Additional authors and addresses can be added with ``\and'',
% just like the second author.
% To save space, use either the email address or home page, not both
}

% conference papers do not typically use \thanks and this command
% is locked out in conference mode. If really needed, such as for
% the acknowledgment of grants, issue a \IEEEoverridecommandlockouts
% after \documentclass

% for over three affiliations, or if they all won't fit within the width
% of the page, use this alternative format:
% 
%\author{\IEEEauthorblockN{Michael Shell\IEEEauthorrefmark{1},
%Homer Simpson\IEEEauthorrefmark{2},
%James Kirk\IEEEauthorrefmark{3}, 
%Montgomery Scott\IEEEauthorrefmark{3} and
%Eldon Tyrell\IEEEauthorrefmark{4}}
%\IEEEauthorblockA{\IEEEauthorrefmark{1}School of Electrical and Computer Engineering\\
%Georgia Institute of Technology,
%Atlanta, Georgia 30332--0250\\ Email: see http://www.michaelshell.org/contact.html}
%\IEEEauthorblockA{\IEEEauthorrefmark{2}Twentieth Century Fox, Springfield, USA\\
%Email: homer@thesimpsons.com}
%\IEEEauthorblockA{\IEEEauthorrefmark{3}Starfleet Academy, San Francisco, California 96678-2391\\
%Telephone: (800) 555--1212, Fax: (888) 555--1212}
%\IEEEauthorblockA{\IEEEauthorrefmark{4}Tyrell Inc., 123 Replicant Street, Los Angeles, California 90210--4321}}

% use for special paper notices
%\IEEEspecialpapernotice{(Invited Paper)}

% make the title area
\maketitle

\begin{abstract}
%\boldmath
Face recognition has been widely studied due to its importance in smart cities applications. However, the case when both training and test images are corrupted is not well solved. To address such a problem, this paper proposes a locality constrained low rank representation and dictionary learning (LCLRRDL) algorithm for robust face recognition. In particular, we present three contributions in the proposed formulation. First, a low-rank representation is introduced to handle the possible contamination of the training as well as test data. Second, a locality constraint is incorporated to acknowledge the intrinsic manifold structure of training data. With the locality constraint term, our scheme induces similar samples to have similar representations. Third, a compact dictionary is learned to handle the problem of corrupted data. The experimental results on two public databases demonstrate the effectiveness of the proposed approach. Matlab code of our proposed LCLRRDL can be downloaded from https://github.com/yinhefeng/LCLRRDL.
\end{abstract}
% IEEEtran.cls defaults to using nonbold math in the Abstract.
% This preserves the distinction between vectors and scalars. However,
% if the journal you are submitting to favors bold math in the abstract,
% then you can use LaTeX's standard command \boldmath at the very start
% of the abstract to achieve this. Many IEEE journals frown on math
% in the abstract anyway.

% Note that keywords are not normally used for peerreview papers.

% For peer review papers, you can put extra information on the cover
% page as needed:
% \ifCLASSOPTIONpeerreview
% \begin{center} \bfseries EDICS Category: 3-BBND \end{center}
% \fi
%
% For peerreview papers, this IEEEtran command inserts a page break and
% creates the second title. It will be ignored for other modes.
\IEEEpeerreviewmaketitle

\section{Introduction}
%\blindtext
The emerging concept of smart city has attracted considerable attention in the urban development policy field and academic research. The concept proposes that smart cities are defined by their ability to solve problems with the help of innovation and Information Communication Technologies (ICTs). Among these technologies, face recognition (FR) has been widely studied due to its wide applications in identity card systems, access control and security monitoring. While in controlled scenarios FR is a solved problem, the performance of unconstrained FR does not meet the requirements of the more challenging applications. Hence, enhancements to FR technologies to support the broad range of applications are  required as a contribution  to the smart city development effort. 

For FR, conventional subspace learning based approaches~\cite{turk1991eigenfaces,belhumeur1997eigenfaces,he2005face,xiao2004new,zheng2006nearest} be applied to reduce the dimension of the face images. Subsequently, the derived low-dimensional features are fed into classifiers to obtain the final recognition result. However, these techniques are not robust to outliers and their performance will be degraded dramatically when test samples are corrupted due to occlusion or disguise. Recently, sparse representation based classification (SRC)~\cite{wright2008robust} has received a lot of attention owing to its impressive robustness to occlusion and other degradations. SRC represents each test image as a sparse linear combination of training image data by solving an $\ell_1$-minimization problem. The classification is then performed by checking which class yields the lowest reconstruction error. Unlike traditional approaches such as Eigenface and Fisherface, SRC does not need an explicit feature extraction phase. The superior performance reported in~\cite{wright2008robust} suggests that it is a promising direction for FR.

Though SRC and its variants are robust to test samples with occlusion and corruption, the performance of them might be degraded when some training and test samples corrupted. Using low-rank matrix recovery (LRMR) for denoising has attracted much attention recently. In particular, Candes \etal\cite{candes2011robust} introduced the robust PCA (RPCA) method, which aims to recover a low-rank matrix from corrupted observations. Furthermore, Liu \etal\cite{liu2012robust} proposed low-rank representation (LRR) method. Based on LRMR, many approaches are presented for robust FR. Yin \etal\cite{yin2016face} presented a new method called low rank matrix recovery with structural incoherence and low rank projection (LRSI\_LRP) which can correct the corrupted test images with a low rank projection matrix. Zhao \etal\cite{zhao2015collaborative} developed a discriminative low-rank representation method for collaborative representation-based (DLRR-CR) robust face recognition. Chen \etal\cite{chen2018robust} proposed a robust low-rank recovery algorithm (RLRR) with a distance-measure structure for face recognition. Li \etal\cite{li2014learning} designed a semi-supervised framework to learn robust face representation with classwise block-diagonal structure. Zhou \etal\cite{zhou2015integrated} presented a supervised low-rank-based approach for learning discriminative features, in which latent low-rank representation (LatLRR) is integrated with a ridge regression-based classifier. Gao \etal\cite{gao2017learning} proposed a novel method, exploiting the low-rank characteristic of both, the data representation and each occlusion-induced error image, allowing the global structure of data together with the error images to be captured simultaneously.

Approaches based on LRMR can effectively deal with the situation that both the training and test samples might be corrupted. However, they either ignore the relationship between similar samples, or they do not learn a compact dictionary from the corrupted training data. To address the above problems, we propose a locality constrained low rank representation and dictionary learning (LCLRRDL) algorithm for robust face recognition. Locality constrained term is introduced to exploit the intrinsic manifold structure of training data. With the locality constraint term, similar samples tend to have similar representations. In contrast to the prior work~\cite{chen2012low,rong2017low} on classification that performs low-rank recovery class by class during training, our method processes all training data simultaneously. Compared to other dictionary learning methods~\cite{akhtar2016discriminative,wang2017cross} that are very sensitive to noise in training images, our dictionary learning algorithm is robust. Contaminated images can be recovered during our dictionary learning process. The main contributions of this paper are summarized as follows.
\begin{itemize}
\item We propose an approach to learning a robust face representation with locality constraint. By incorporating the locality constrained term, this approach encourages similar samples to have similar representations. The learned representation can be employed for classification directly.
\item A compact dictionary with better reconstruction and discrimination capability can be learned by our approach.
\item Our approach obtains the representations for all training samples simultaneously in a computationally efficient manner.
\end{itemize}

This paper is organized as follows: Section II presents related works on low rank matrix recovery. Detailed description of the proposed method is presented in Section III. An efficient optimization algorithm based on the inexact Augmented Lagrange Multiplier method is presented in Section IV. Experiments conducted on two public databases are given in Section V. Finally, Section VI concludes the paper.

\section{Related Work}
\label{Sec_2}
Robust PCA~\cite{candes2011robust} is a representative work of LRMR. It seeks to decompose corrupted observations into two matrices, one is a low-rank matrix and the other is the associated sparse error matrix. More specifically, to derive the low-rank approximation of the input data matrix $\mathbf{X}$, RPCA minimizes the rank of matrix $\mathbf{A}$ while reducing the $\ell_0$-norm of $\mathbf{E}$. As the optimization of rank function and $\ell_0$-norm is highly nonconvex, we can get the following tractable convex optimization surrogate by replacing the rank function with nuclear norm $\left \| \mathbf{A} \right \|_*$(\ie, the sum of the singular values of $\mathbf{A}$), and the $\ell_0$-norm with the $\ell_1$-norm,
\begin{equation}
\label{eq:obj_rpca}
\underset{\mathbf{A},\mathbf{E}}{\textrm{min}}\,\left \| \mathbf{A}\right \|_{*}+\beta \left \| \mathbf{E} \right \|_{1} , \ \textrm{s.t.} \ \mathbf{X}=\mathbf{A}+\mathbf{E}
\end{equation}
RPCA implicitly assumes that the underlying data structure is a single low-rank subspace. However, a more reasonable assumption is that the data samples are approximately drawn from a union of multiple subspaces. Recently, Liu \etal\cite{liu2012robust} generalized the concept of RPCA and presented a more general rank minimization problem, defined as follows,

\begin{equation}
\label{eq:obj_lrr}
\underset{\mathbf{Z},\mathbf{E}}{\textrm{min}}\,\left \| \mathbf{Z}\right \|_{*}+\beta \left \| \mathbf{E} \right \|_{l} , \ \textrm{s.t.} \ \mathbf{X}=\mathbf{D}\mathbf{Z}+\mathbf{E}
\end{equation}
where $\left \| . \right \|_l$ indicates a certain regularization strategy for characterizing various corruptions.

\section{Our proposed method}
\label{Sec_3}
\subsection{Locality Constrained Low Rank Representation}
The importance of the geometrical information of samples for discrimination has been shown in many papers. Motivated by the recent progress in manifold learning, in this paper we employ a locality constrained term which explicitly considers the geometrical structure of samples. Most of the manifold learning algorithms use the locally invariant idea that if two data points $\boldsymbol{x}_i$ and $\boldsymbol{x}_j$ are close in the intrinsic structure of the data distribution, then they will have a large weight $\mathbf{R}_{ij}$ between the two points and are likely to exhibit affinity in the new data representation space.

As in~\cite{wei2015latent}, we utilize the following locality constrained term to discover the local geometrical information,
\begin{equation}
\label{eq:weight_term}
\left \| \mathbf{R}\odot \mathbf{Z} \right \|_1
\end{equation}
where $\odot$ denotes the Hadamard product and $\mathbf{R}_{ij}=\left \| \mathbf{x}_i-\mathbf{x}_j \right \|_2^2$. (\ref{eq:weight_term}) can be regarded as a weighted  $\ell_1$-norm, thus it can promote the sparsity of the objective function. According to~\cite{belkin2003laplacian}, it is beneficial for classification when using a sparse graph which characterizes the locality relationships.

The weight matrix $\mathbf{R}$ has a beneficial property: the smaller the weight is, the more similar the samples are, while the larger the weight is, the more different the samples are. As such, it is expected to achieve better intraclass compactness and interclass separation.

With $\mathbf{R}$ in hand, the local constraint term is integrated into the objective function of LRR to yield
\begin{equation}
\label{eq:obj_lclrr}
\underset{\mathbf{Z},\mathbf{E}}{\textrm{min}} \ \left \| \mathbf{Z} \right \|_*+\lambda\left \| \mathbf{E} \right \|_{21}+\alpha \left \| \mathbf{R}\odot \mathbf{Z} \right \|_1, \ \textrm{s.t.} \ \mathbf{X}=\mathbf{D}\mathbf{Z}+\mathbf{E}
\end{equation}
Here the $\ell_{21}$-norm of $\mathbf{E}$ is used to model sample-specific corruptions and outliers. This model will be called locality constrained low-rank representation (LCLRR).

\subsection{Dictionary Learning}
Dictionary quality is of great importance for the face recognition problem, especially for the case that both training and testing images are corrupted~\cite{zhang2013learning}. The performance of the classification algorithm has been improved significantly with learned dictionary from the corrupted training data~\cite{chen2012low,rong2017low}. Additionally, the representation learning process becomes efficient with respect to a compact dictionary, other than utilizing the whole training data as dictionary~\cite{wright2008robust,liu2012robust}. Here, we attempt to learn compact and discriminative dictionary from corrupted observations by exploiting the local geometrical information. The objective function is proposed to learn robust representation and dictionary simultaneously with locality constraint term as follows,
\begin{equation}
\label{eq:obj_pro}
\underset{\mathbf{Z},\mathbf{E},\mathbf{D}}{\textrm{min}} \ \left \| \mathbf{Z} \right \|_*+\lambda\left \| \mathbf{E} \right \|_{21}+\alpha \left \| \mathbf{R}\odot \mathbf{Z} \right \|_1+\frac{\gamma}{2}\left \| \mathbf{D} \right \|_F^2, \ \textrm{s.t.} \ \mathbf{X}=\mathbf{D}\mathbf{Z}+\mathbf{E}
\end{equation}
where $\frac{\gamma}{2}\left \| \mathbf{D} \right \|_F^2$ is to avoid scale change during the dictionary learning process.

\subsection{Classification}
We use a linear classifier for classification. When (\ref{eq:obj_pro}) is solved, the low-rank representations $\mathbf{Z}$ of training data $\mathbf{X}$ can be obtained. For test data $\mathbf{X}_{test}$, their representations $\mathbf{Z}_{test}$ can be derived by solving (\ref{eq:obj_lrr}) with the dictionary learned from (\ref{eq:obj_pro}). The representation $\boldsymbol{z}_i$ for test sample $i$ is the $i$th column vector in $\mathbf{Z}_{test}$. We use the multivariate ridge regression model to obtain a linear classifier $\mathbf{W}$ :
\begin{equation}
\label{eq:obj_classifier}
\mathbf{W}=\textrm{arg} \ \underset{\mathbf{W}}{\textrm{min}} \ \left \| \mathbf{H}-\mathbf{W}\mathbf{Z} \right \|+\eta\left \| \mathbf{W} \right \|_F^2
\end{equation}
where $\mathbf{H}$ is the class label matrix of $\mathbf{X}$. This yields
\begin{equation}
\label{eq:solu_classifier}
\mathbf{W}=\mathbf{H}\mathbf{Z}^T(\mathbf{Z}\mathbf{Z}^T+\eta \mathbf{I})^{-1}
\end{equation}
Then label for sample $i$ is given by
\begin{equation}
\label{eq:rule_classify}
k=\textrm{arg} \ \underset{k}{\textrm{max}} \ \mathbf{W}\mathbf{z}_i
\end{equation}
where $k$ is corresponding to the classifier with the largest output.

\section{Optimization}
\label{Sec_4}
To solve optimization problem~(\ref{eq:obj_pro}), we first introduce auxiliary variables $\mathbf{J}$ and $\mathbf{L}$ to make the objective function separable. Problem ~(\ref{eq:obj_pro}) can be reformulated as:
\begin{equation}
\label{eq:equ_formu}
\begin{split}
&\underset{\mathbf{Z},\mathbf{J},\mathbf{L},\mathbf{E},\mathbf{D}}{\textrm{min}} \ \left \| \mathbf{J} \right \|_*+\lambda\left \| \mathbf{E} \right \|_{21}+\alpha \left \| \mathbf{R}\odot \mathbf{L} \right \|_1+\frac{\gamma}{2}\left \| \mathbf{D} \right \|_F^2 \\ &\textrm{s.t.}  \  \mathbf{X}=\mathbf{D}\mathbf{Z}+\mathbf{E},\mathbf{Z}=\mathbf{J},\mathbf{Z}=\mathbf{L}
\end{split}
\end{equation}
which can be solved based on the Augmented Lagrange Multiplier (ALM) method~\cite{lin2010augmented}. The augmented Lagrangian function of (\ref{eq:equ_formu}) is:
\begin{equation}
\label{eq:augment}
\begin{split}
&\mathcal{L} (\mathbf{Z},\mathbf{J},\mathbf{L},\mathbf{E},\mathbf{D},\mathbf{Y}_1,\mathbf{Y}_2,\mathbf{Y}_3,\mu)= \left \| \mathbf{J} \right \|_*+\lambda\left \| \mathbf{E} \right \|_{21}\\&+\alpha \left \| \mathbf{R}\odot \mathbf{L} \right \|_1+ \frac{\gamma}{2}\left \| \mathbf{D} \right \|_F^2+<\mathbf{Y}_1,\mathbf{X}-\mathbf{D}\mathbf{Z}-\mathbf{E}>\\&+<\mathbf{Y}_2,\mathbf{Z}-\mathbf{J}>+<\mathbf{Y}_3,\mathbf{Z}-\mathbf{L}> \\ &+\frac{\mu}{2}(\left \| \mathbf{X}-\mathbf{D}\mathbf{Z}-\mathbf{E} \right \|_F^2+\left \| \mathbf{Z}-\mathbf{J} \right \|_F^2+\left \| \mathbf{Z}-\mathbf{L} \right \|_F^2)
\end{split}
\end{equation}
where $\left \langle \mathbf{A},\mathbf{B} \right \rangle=trace(\mathbf{A}^t\mathbf{B})$. $\mathbf{Y}_1$, $\mathbf{Y}_2$ and $\mathbf{Y}_3$ are Lagrange multipliers and $\mu>0$ is a penalty parameter. The optimization of (\ref{eq:augment}) can be solved iteratively by updating $\mathbf{J}$, $\mathbf{Z}$, $\mathbf{L}$, $\mathbf{E}$ and $\mathbf{D}$ one at a time. The updating scheme is as follows.

Updating $\mathbf{J}$: Fix the other variables and solve the following problem, 
\begin{equation}
\label{eq:update_J}
\begin{split}
&\mathbf{J}^{k+1}=\textrm{arg} \ \underset{\mathbf{J}}{\textrm{min}} \ \left \| \mathbf{J} \right \|_*+<\mathbf{Y}_2^k,\mathbf{Z}^k-\mathbf{J}>+\frac{\mu^k}{2}\left \| \mathbf{Z}^k-\mathbf{J} \right \|_F^2 \\ &=\textrm{arg} \ \underset{\mathbf{J}}{\textrm{min}} \ \frac{1}{\mu^k}\left \| \mathbf{J} \right \|_*+\frac{1}{2}\left \| \mathbf{J}-(\mathbf{Z}^k+\frac{\mathbf{Y}_2^k}{\mu^k}) \right \|_F^2
 \\ &=\mathbf{U}S_{\frac{1}{\mu^k}}[\mathbf{\Sigma} ]\mathbf{V}^T
\end{split}
\end{equation}
where $(\mathbf{U},\mathbf{\Sigma},\mathbf{V}^T)=SVD(\mathbf{Z}^k+\mathbf{Y}_2^k/\mu^k)$ and $S_{\varepsilon }[.] $ is the shrinkage operator defined as follows~\cite{lin2010augmented},
\begin{equation}
\label{eq:soft_thres}
S_{\varepsilon }[x]=\left\{\begin{matrix}
x-\varepsilon, \textrm{if}  \ x>\varepsilon\\ 
x+\varepsilon, \textrm{if}  \ x<-\varepsilon\\ 
0,\textrm{otherwise}
\end{matrix}\right. 
\end{equation}
Updating $\mathbf{Z}$: Fix the other variables and solve the following problem,
\begin{equation}
\label{eq:update_Z}
\begin{split}
&\mathbf{Z}^{k+1}=\textrm{arg} \ \underset{\mathbf{Z}}{\textrm{min}} \ <\mathbf{Y}_1^k,\mathbf{X}-\mathbf{D}^k\mathbf{Z}-\mathbf{E}^k>\\&+<\mathbf{Y}_2^k,\mathbf{Z}-\mathbf{J}^{k+1}>+<\mathbf{Y}_3^k,\mathbf{Z}-\mathbf{L}^k>\\&+\frac{\mu^k}{2}(\left \| \mathbf{X}-\mathbf{D}^k\mathbf{Z}-\mathbf{E}^k \right \|_F^2+\left \| \mathbf{Z}-\mathbf{J}^{k+1} \right \|_F^2+\left \| \mathbf{Z}-\mathbf{L}^k \right \|_F^2)
\end{split}
\end{equation}
(\ref{eq:update_Z}) has the following closed-form solution,
\begin{equation}
\label{eq:solu_Z}
\begin{split}
&\mathbf{Z}^{k+1}=[(\mathbf{D}^k)^T\mathbf{D}^k+2\mathbf{I}]^{-1}[(\mathbf{D}^k)^T(\mathbf{X}-\mathbf{E}^k)+\mathbf{J}^{k+1}+\mathbf{L}^k\\&+\frac{(\mathbf{D}^k)^T\mathbf{Y}_1^k-\mathbf{Y}_2^k-\mathbf{Y}_3^k}{\mu^k}]
\end{split}
\end{equation}
Updating $\mathbf{L}$: Fix the other variables and solve the following problem,
\begin{equation}
\label{eq:update_L}
\begin{split}
&\mathbf{L}^{k+1}=\textrm{arg} \ \underset{\mathbf{L}}{\textrm{min}} \ \alpha \left \| \mathbf{R}\odot \mathbf{L} \right \|_1+<\mathbf{Y}_3^k,\mathbf{Z}^{k+1}-\mathbf{L}>\\&+\frac{\mu^k}{2}\left \| \mathbf{Z}^{k+1}-\mathbf{L} \right \|_F^2
\\&=\textrm{arg} \ \underset{\mathbf{L}}{\textrm{min}} \ \alpha \left \| \mathbf{R}\odot \mathbf{L} \right \|_1+\frac{\mu^k}{2}\left \| \mathbf{L}-(\mathbf{Z}^{k+1}+\frac{\mathbf{Y}_3^k}{\mu^k}) \right \|_F^2
\end{split}
\end{equation}
which can be updated by the elementwise strategy. For the $i$th row and $j$th column element $\mathbf{L}_{ij}$, the optimal solution of problem (\ref{eq:update_L}) is
\begin{equation}
\label{eq:solu_L}
\begin{split}
&\mathbf{L}_{ij}^{k+1}=\textrm{arg} \ \underset{\mathbf{L_{ij}}}{\textrm{min}} \ \alpha \mathbf{R}_{ij}\left | \mathbf{L}_{ij} \right |+\frac{\mu^k}{2}\left \| \mathbf{L}_{ij}-\mathbf{M}_{ij} \right \|_F^2\\&
=S_{\frac{\alpha \mathbf{R}_{ij}}{\mu^k}}(\mathbf{M}_{ij})
\end{split}
\end{equation}
where $\mathbf{M}_{ij}=\mathbf{Z}_{ij}^{k+1}+((\mathbf{Y}_3^{k})_{ij}/\mu^k) $.

Updating $\mathbf{E}$: Fix the other variables and solve the following problem,
\begin{equation}
\label{eq:update_E}
\begin{split}
&\mathbf{E}^{k+1}=\textrm{arg} \ \underset{\mathbf{E}}{\textrm{min}} \ \lambda \left \| \mathbf{E} \right \|_{21}+<\mathbf{Y}_1^k,\mathbf{X}-\mathbf{D}^k\mathbf{Z}^{k+1}-\mathbf{E}>\\&+\frac{\mu^k}{2}\left \| \mathbf{X}-\mathbf{D}^k\mathbf{Z}^{k+1}-\mathbf{E} \right \|_F^2\\&=\textrm{arg} \ \underset{\mathbf{E}}{\textrm{min}} \ \frac{\lambda}{\mu^k}\left \| \mathbf{E} \right \|_{21}+\frac{1}{2}\left \| \mathbf{E}-(\mathbf{X}-\mathbf{D}^k\mathbf{Z}^{k+1}+\frac{\mathbf{Y}_1^k}{\mu^k}) \right \|_F^2
\end{split}
\end{equation}

Updating $\mathbf{D}$: Fix the other variables and solve the following problem,
\begin{equation}
\label{eq:solu_D}
\begin{split}
&\mathbf{D}^{k+1}=\textrm{arg} \ \underset{\mathbf{D}}{\textrm{min}} \ \frac{\gamma}{2}\left \| \mathbf{D} \right \|_F^2+<\mathbf{Y}_1,\mathbf{X}-\mathbf{D}\mathbf{Z}^{k+1}-\mathbf{E}^{k+1}>\\&+\frac{\mu^k}{2}\left \| \mathbf{X}-\mathbf{D}\mathbf{Z}^{k+1}-\mathbf{E}^{k+1} \right \|_F^2\\
&=[\frac{\mathbf{Y}_1^k(\mathbf{Z}^{k+1})^T}{\mu^k}-(\mathbf{E}^{k+1}-\mathbf{X})(\mathbf{Z}^{k+1})^T](\frac{\gamma}{\mu^k}\mathbf{I}+\mathbf{Z}^{k+1}(\mathbf{Z}^{k+1})^T)^{-1}
\end{split}
\end{equation}
Algorithm 1 summarizes the solution to problem (\ref{eq:equ_formu}).

\begin{algorithm} 
\caption{Solving Problem (\ref{eq:equ_formu}) by Inexact ALM} 
\label{alg1} 
\begin{algorithmic}[1]
\REQUIRE Training data matrix $\mathbf{X}$; Parameters $\lambda$, $\alpha$ and $\gamma$; Weight matrix $\mathbf{R}$
\STATE Initialize $\mathbf{Z}^0=0$, $\mathbf{J}^0=0$, $\mathbf{L}^0=0$, $\mathbf{E}^0=0$, $\mathbf{Y}_1^0=0$, $\mathbf{Y}_2^0=0$, $\mathbf{Y}_3^0=0$, $\mu_{max}=10^8$, $\varepsilon=10^{-6}$, $\rho=1.15$
\WHILE{not converged} 
\STATE fix the others and update $\mathbf{J}$ by~(\ref{eq:update_J})
\STATE fix the others and update $\mathbf{Z}$ by~(\ref{eq:solu_Z})
\STATE fix the others and update $\mathbf{L}$ by~(\ref{eq:solu_L})
\STATE fix the others and update $\mathbf{E}$ by~(\ref{eq:update_E})
\STATE fix the others and update $\mathbf{D}$ by~(\ref{eq:solu_D})
\STATE update the multipliers:

$\mathbf{Y}_1^{k+1}=\mathbf{Y}_1^{k}+\mu^k(\mathbf{X}-\mathbf{D}^{k+1}\mathbf{Z}^{k+1}-\mathbf{E}^{k+1})$

$\mathbf{Y}_2^{k+1}=\mathbf{Y}_2^{k}+\mu^k(\mathbf{Z}^{k+1}-\mathbf{J}^{k+1})$

$\mathbf{Y}_3^{k+1}=\mathbf{Y}_3^{k}+\mu^k(\mathbf{Z}^{k+1}-\mathbf{L}^{k+1})$

\STATE update $\mu$:

$\mu^{k+1}=\textrm{min}(\mu_{max},\rho\mu^k)$

\STATE check the convergence conditions

$\left \| \mathbf{Z}^{k+1}-\mathbf{J}^{k+1} \right \|_{\infty}<\varepsilon$, $\left \| \mathbf{Z}^{k+1}-\mathbf{L}^{k+1} \right \|_{\infty}<\varepsilon$ and $\left \| \mathbf{X}-\mathbf{D}^{k+1}\mathbf{Z}^{k+1}-\mathbf{E}^{k+1} \right \|_{\infty}<\varepsilon$
\ENDWHILE 
\ENSURE $\mathbf{Z}$, $\mathbf{D}$ and $\mathbf{E}$
\end{algorithmic} 
\end{algorithm}

\section{Experimental Evaluation}
\label{Sec_4}
To demonstrate the efficacy of our proposed LCLRRDL, experiments are conducted on two publicly available databases: the Extended Yale B database~\cite{georghiades2001few} and the AR database~\cite{martinez2001pca}. We tackle the scenario that both training and testing images are corrupted due to illumination changes, expression changes, pose changes and occlusion. Our algorithm (LCLRRDL) is compared with related approaches including LLC~\cite{wang2010locality}, SRC~\cite{wright2008robust}, RPCA~\cite{candes2011robust}, LRSI~\cite{chen2012low}, SLRR~\cite{zhang2013learning}, and LRSR~\cite{li2014learning}. Here SRW indicates the case that the whole training set are used as the dictionary, SRS indicates the case that the dictionary size is the same as ours.
\subsection{Experiments on the Extended Yale B Database}
\begin{figure}[!t]
\centering
\includegraphics[width=2.5in]{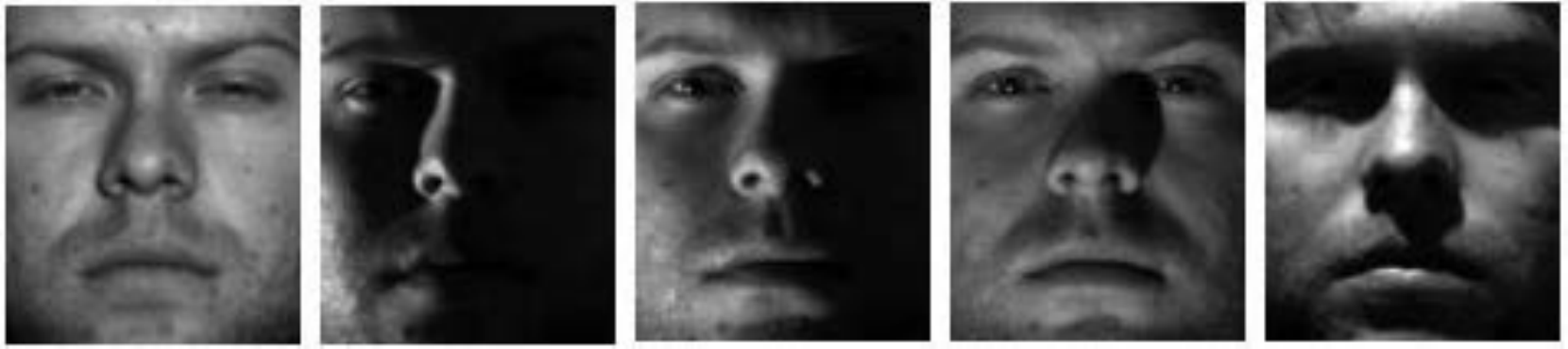}
\caption{Example images from the Extended Yale B database.}
\label{fig:exam_eyaleb}
\end{figure}

The Extended Yale B database contains 2414 frontal-face images of 38 individuals (about 64 images per subject). The images are taken under different poses and illumination conditions with size 192$\times$168, Fig.~\ref{fig:exam_eyaleb} shows several example images from this database. We test our approach on the down-sampled images with sample rate 1/2, 1/4, 1/8, and the corresponding feature dimension is 8064, 2016 and 504. Following the protocol in SLRR~\cite{zhang2013learning}, we first randomly select $N_c$ images for each person as training images ($N_c$ = 8, 32), and the rest as testing images. For the case of 8 training images per class, a compact dictionary with 5 items for each class is preferable. For the case of 32 training images per class, a compact dictionary with 20 items for each class is preferred. Comparison of different methods on the Extended Yale B database is presented in Table~\ref{table:EYaleB}. We can find that our approach (LCLRRDL) outperforms its competitors in both cases. It outperforms SLRR by 1.8\% improvement with 8 training images per person, and 5.8\% improvement with 32 training images per person on average. Our approach achieves significant performance gains in the case of 32 training images per person, since dictionary learned with 20 items for each class has better capability to expand corresponding subspace, and the representation derived by LCLRRDL has better discrimination capability, which is beneficial for classification purpose.

\begin{table}[t]
\centering
\caption{Recognition accuracy in (\%) on the Extended Yale B database}
\label{table:EYaleB}
\begin{tabular}{ccccccc}
\hline
No. per class & \multicolumn{3}{c}{$N_c$=8}                      & \multicolumn{3}{c}{$N_c$=32}                     \\ \hline
Sample Rate   & 1/8           & 1/4           & 1/2           & 1/8           & 1/4           & 1/2           \\ \hline
\textbf{LCLRRDL} & \textbf{80.1} & \textbf{84.6} & \textbf{85.0} & \textbf{97.8} & \textbf{99.0} & \textbf{99.8} \\
LRSR~\cite{li2014learning}          & 75.3          & 78.6          & 79.5          & 96.8          & 96.9          & 97.7          \\
SLRR~\cite{zhang2013learning}          & 76.6          & 83.7          & 83.8          & 89.9          & 93.6          & 95.7          \\
LRSI~\cite{chen2012low}          & 73.3          & 80.9          & 80.8          & 89.5          & 93.5          & 94.5          \\
RPCA~\cite{candes2011robust}          & 74.6          & 78.3          & 80.2          & 85.6          & 90.7          & 94.1          \\
SRW~\cite{wright2008robust}           & 79.3          & 83.0          & 83.8          & 87.2          & 89.5          & 90.7          \\
SRS~\cite{wright2008robust}           & 75.3          & 78.9          & 80.1          & 84.4          & 85.7          & 85.9          \\ 
LLC~\cite{wang2010locality}           & 65.7          & 70.6          & 76.1          & 76.4          & 80.0          & 85.6          \\ \hline
\end{tabular}
\end{table}

\subsection{Experiments on the AR Database}
\begin{figure}[!t]
\centering
\includegraphics[width=2.5in]{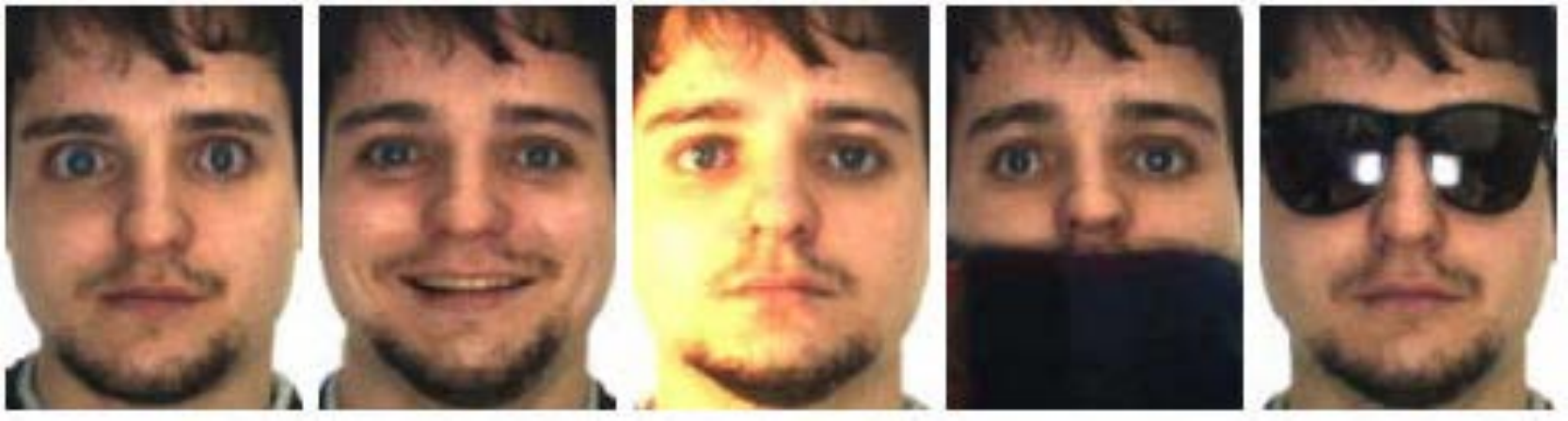}
\caption{Example images from the AR database.}
\label{fig:exam_ar}
\end{figure}

The AR database consists of over 4000 images corresponding to 126 individuals. For each individual, there are 26 images taken in two separate sessions under different illumination and expression changes. For each session, there are thirteen images, among which three images are with sunglasses, another three with scarves, and the remaining seven images exhibit different illumination and expression variations (and thus are considered as clean/neutral images). Fig.~\ref{fig:exam_ar} shows several example images from this database, the size of each image is 165$\times$120 pixels. In our experiments, we choose a subset of the AR database consisting of 50 men and 50 women. Following the protocol in SLRR~\cite{zhang2013learning}, we convert the color images to gray scale and down-sample them at the rate of 1/3. The dimension of the resulting feature vector is 2200. Experiments are conducted in the following three scenarios:

1) Sunglasses: We first consider occluded training samples due to the presence of sunglasses, which affect about 20\% of the face image. We use seven neutral images plus one image with sunglasses (randomly chosen) from session 1 for training (eight training images per class), and the remaining neutral images (all from session 2) and the rest of the images with sunglasses (two taken at session 1 and three at session 2) for testing (twelve test images per class). 

2) Scarf: Replace images with sunglasses in the above scenario by images with scarf.

3) Mixed (Sunglasses and Scarf): In the last scenario, the training samples are occluded by sunglasses and scarves, which is more challenging than the above two scenarios. Seven neutral  images, two corrupted images (one with sunglasses and one with scarf) from session 1 are used for training (nine training images per class), and the rest are used for testing (seventeen test images per class).

As in SLRR~\cite{zhang2013learning}, a compact dictionary with 5 items for each class is preferable in all different scenarios. A comparison of different methods on the AR database is summarized in Table~\ref{table:AR}. Our approach achieves the best recognition result and outperforms SLRR by 5.0\% for the sunglasses scenario, 7.0\% for the scarf scenario, and 7.6\% for the mixed scenario. Our approach shows robustness to severe occlusions like sunglasses and scarf. In contrast, the performance of LRSR and LRSI is not satisfactory when both training and test images are badly corrupted. High quality dictionary is critical for learning discriminative representation when both training and test images are badly corrupted.

\begin{table}[t]
\centering
\caption{Recognition accuracy (\%) on the AR database}
\label{table:AR}
\begin{tabular}{cccc}
\hline
Scenario      & Sunglasses    & Scarf         & Mixed         \\ \hline
\textbf{LCLRRDL} & \textbf{92.3} & \textbf{90.4} & \textbf{90.0} \\
LRSR~\cite{li2014learning}          & 89.2          & 85.2          & 85.6          \\
SLRR~\cite{zhang2013learning}          & 87.3          & 83.4          & 82.4          \\
LRSI~\cite{chen2012low}          & 84.9          & 76.4          & 80.3          \\
RPCA~\cite{candes2011robust}          & 83.2          & 75.8          & 78.9          \\
SRW~\cite{wright2008robust}           & 86.8          & 83.2          & 79.2          \\
SRS~\cite{wright2008robust}           & 82.1          & 72.6          & 65.5          \\ 
LLC~\cite{wang2010locality}           & 65.3          & 59.2          & 59.9          \\ \hline
\end{tabular}
\end{table}

\subsection{Evaluation of classification time}
We conducted experiments on the AR dataset in the sunglasses scenario to compare the training time of the proposed LCLRRDL algorithm with the benchmark methods. The training time refers to the time taken to obtain the low rank representation and (or) compact dictionary from the training data. The experimental settings are the same as in the previous subsection. Table~\ref{table:train_time} shows the training time of all the methods on the AR dataset. Since LLC, SRS and SRW do not involve low rank representation, here we just list the training time of the remaining approaches. 

\begin{table}[t]
\centering
\caption{Training time of different methods}
\label{table:train_time}
\begin{tabular}{cc}
\hline
Methods      & Training time (s)            \\ \hline
\textbf{LCLRRDL} & 38.7  \\
LRSR~\cite{li2014learning}          & 323.2               \\
SLRR~\cite{zhang2013learning}          & 98.2         \\
LRSI~\cite{chen2012low}          & 2.1          \\
RPCA~\cite{candes2011robust}           & 7.4        \\ \hline
\end{tabular}
\end{table}

All the experiments are conducted on a 64-bit computer with Inter i7-4790 3.6 GHz CPU and 16GB RAM under the MATLAB R2016b programming environment. As shown in Table~\ref{table:train_time}, RPCA and LRSI cost less time than the other methods due to their simple objective function. The training time of LCLRRDL is less than that of SLRR and LRSR, which indicates that it can be applied in practical scenarios.

\section{Conclusion}
\label{Sec_5}
%\blindtext
In this paper, we propose a locality constrained low rank representation to learn robust representation for face recognition in the case of both training and test images being corrupted. By introducing the locality constraint term into the framework, the proposed approach can fully exploit the intrinsic manifold structure of the training data. As part of the representation learning process, a compact dictionary with better reconstruction capability is learned by our approach. The proposed robust representation provides input to a simple, yet powerful, linear multi-classifier designed for the final classification task. The experimental results on benchmark databases demonstrate the efficacy of the proposed approach.
% if have a single appendix:
%\appendix[Proof of the Zonklar Equations]
% or
%\appendix  % for no appendix heading
% do not use \section anymore after \appendix, only \section*
% is possibly needed

% use appendices with more than one appendix
% then use \section to start each appendix
% you must declare a \section before using any
% \subsection or using \label (\appendices by itself
% starts a section numbered zero.)
%

%\appendices
%\section{Proof of the First Zonklar Equation}
%\blindtext

% use section* for acknowledgement
\section*{Acknowledgements}
The work was supported by the National Natural Science Foundation of China (61672265,U1836218), the 111 Project of the Ministry of Education of China (B12018), the Postgraduate Research and Practice Innovation Program of Jiangsu Province under Grant No. KYLX\_1123, the Overseas Studies Program for Postgraduates of Jiangnan University and the China Scholarship Council (CSC, No.201706790096), the EPSRC programme grant (FACER2VM) under the number EP/N007743/1, the U.S. Army Research Laboratory, the U. S. Army Research Office, the U.K. Ministry of Defence and the U.K. EPSRC grant under the number EP/R013616/1.

{\small
\bibliographystyle{IEEEtran}
\bibliography{mybib}
}

\end{document}